\documentclass[runningheads]{llncs}
\usepackage{graphicx}
\usepackage{amsmath,amssymb} %
\usepackage{color}
\usepackage{times}
\usepackage{epsfig}
\usepackage{xcolor}
\usepackage{makecell}
\usepackage{multirow}
\usepackage{hyperref}

\definecolor{orange}{rgb}{1,0.5,0}
\definecolor{deeppink}{RGB}{255,20,147}

\newif\ifdraft

\draftfalse
\begin{document}
	
\pagestyle{headings}
\mainmatter

\title{3D Pose Based Feedback For Physical Exercises} %
\titlerunning{3D Pose Based Feedback For Physical Exercises}
\authorrunning{Z. Zhao et al.}

\author{Ziyi Zhao\inst{1}, Sena Kiciroglu\inst{1}, Hugues Vinzant\inst{1}, \\Yuan Cheng\inst{1}, Isinsu Katircioglu\inst{1}, Mathieu Salzmann\inst{1,2}, Pascal Fua\inst{1}}
\institute{CVLab EPFL, Switzerland \and ClearSpace, Switzerland}
\maketitle

Unsupervised self-rehabilitation exercises and physical training can cause serious injuries if performed incorrectly. We introduce a learning-based framework that identifies the mistakes made by a user and proposes corrective measures for easier and safer individual training. Our framework does not rely on hard-coded, heuristic rules. Instead, it learns them from data, which facilitates its adaptation to specific user needs. To this end, we use a Graph Convolutional Network (GCN) architecture acting on the user's pose sequence to model the relationship between the the body joints trajectories. To evaluate our approach, we introduce a dataset with 3 different physical exercises. Our approach yields $90.9\%$ mistake identification accuracy and successfully corrects $94.2\%$ of the mistakes.

\section{Introduction}
\begin{figure}[ht!]
	\centering
	\includegraphics[width=1.0\linewidth]{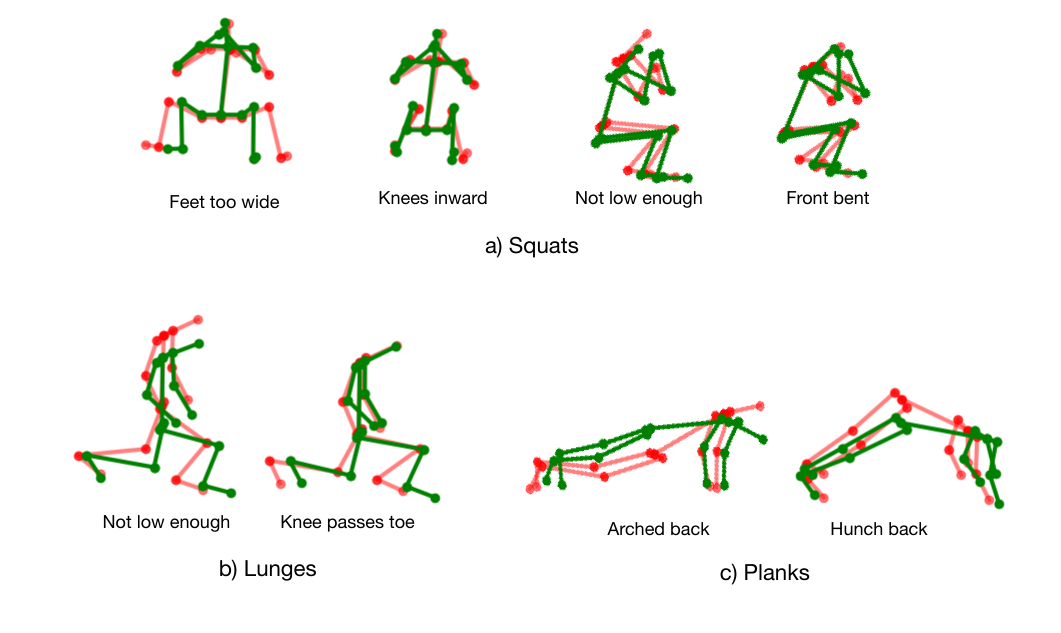}
	\caption{{\bf Example results from our framework} depicting frames from the a) squat, b) lunge, and c) plank classes. The red poses correspond to the exercises performed incorrectly while the green poses correspond to our corrections. Note that although we display a single pose from each mistake type, our framework operates on entire sequences.}
	\label{fig:teaser}
\end{figure}

Being able to perform exercises without requiring the supervision of a physical trainer is a convenience many people enjoy, especially after the COVID-19 pandemic. However, the lack of effective supervision and feedback can end up doing more harm than good, which may include causing serious injuries. There is therefore a growing need for computer-aided exercise feedback strategies. 
A few recent works have addressed this problem~\cite{Chen20d,Yang21,Kanase21,Fieraru21,Dittakavi22,Rangari22}. However, they focus only on identifying whether an exercise is performed correctly or not~\cite{Chen20d,Rangari22}, or they rely on hard-coded rules based on joint angles that cannot easily be extended to new exercises~\cite{Yang21,Kanase21,Fieraru21}. In this work, we therefore leverage recent advances in the fields of pose estimation~\cite{Kocabas20,Zheng21,Gong22}, action recognition~\cite{Li19m,zhang20e} and motion prediction~\cite{Aksan21,Mao20,Kiciroglu22} to design a framework that provides automated and personalized feedback to supervise physical exercises.
Specifically, we developed a method that not only points out mistakes but also offers suggestions on how to fix them without relying on hard-coded, heuristic rules to define what a successful exercise sequence should be. Instead, it learns from data. To this end, we use a two-branch deep network. One branch is an action classifier that tells users what kind of errors they are making. The other proposes corrective measures. They both rely on Graph Convolutional Networks (GCNs) that can learn to exploit  the relationships between the trajectories of individual joints. Fig.~\ref{fig:teaser} depicts the kind of output our network produces.

To showcase our framework's performance, we recorded a physical exercise dataset with 3D poses and instruction label annotations. Our dataset features $3$ types of exercises; squats, lunges and planks. Each exercise type is performed correctly and with mistakes following specific instructions by $4$ different subjects. Our approach achieves $90.9\%$ mistake recognition accuracy on a test set. Furthermore, we use the action classification branch of our framework to evaluate the performance of the correction branch, considering the correction to be successful if the corrected motion is classified as ``correct". Under this metric, our approach successfully corrects $94.2\%$ of users' mistakes.
\section{Related Work}
Our work is at the intersection of several sub-fields of computer vision: (i) We draw inspiration from GCN based \textbf{human motion prediction} architectures; (ii) we identify the users' mistakes in an \textbf{action recognition} fashion; and (iii) we address the task of \textbf{physical exercise analysis}. We therefore discuss these three topics below.
\subsection{Human Motion Prediction}
Human motion prediction is a complex task due to the inherent uncertainty in forecasting into the future. In recent years, many deep learning methods based on recurrent neural networks  (RNNs)~\cite{Fragkiadaki15,Jain16,Ghosh17,Martinez17b,Wang19h,Barsoum18,Kiciroglu22}, variational auto encoders (VAEs)~\cite{Butepage17,Butepage18b,Butepage19,Aliakbarian20}, transformers~\cite{Aksan21}, and graph convolutional networks (GCNs)~\cite{Mao19,Lebailly20,Mao20,Katircioglu22} have been proposed. We focus our discussion on GCN based ones, as we also exploit the graph-like connections of human joints with GCNs in our approach. 

In~\cite{Mao19}, Mao \textit{et al.} proposed to exploit a GCN to model the relationships across joint trajectories by representing them with Discrete Cosine Transform (DCT) coefficients. The approach was in \cite{Mao20} by integrating an attention module, and in \cite{Katircioglu22,Guo22} by using cross-subject attention for multi-person motion prediction. In \cite{Lebailly20}, the input coefficients to the GCN were extracted via an inception module instead of the DCT. Our motion correction branch is inspired by~\cite{Mao19}, but instead of forecasting future motion, we predict correctly performed exercises.  
\subsection{Action Recognition}
Although there is a vast literature on image-based action recognition, here we focus on its skeleton-based counterpart, as our approach also processes 3D poses.
Early deep learning based approaches to skeleton-based action recognition mostly relied on RNNs \cite{Du15,shahroudy16,Liu16c,Song17,Liu17}. Li \textit{et al.}~\cite{Li19m} used convolutional neural networks (CNNs) to extract features hierarchically by first finding local point-level features and gradually extracting global spatial and temporal features. Zhang \textit{et al.}~\cite{Zhang19f} designed CNN and RNN networks that are robust to viewpoint changes.  
Recently, \cite{Tang18e,Li19n,zhang20e} employed GCNs for action recognition. Specifically, Tang \textit{et al.}~\cite{Tang18e} designed a reinforcement learning scheme to select the most informative frames and feed them to a GCN. Li \textit{et al.}~\cite{Li19n} developed a GCN framework that not only models human joint connections, but also learns to infer ``actional-links", which are joint dependencies learned from the data. 
Zhang \textit{et al.}~\cite{zhang20e} designed a two-module network, consisting of a first GCN-based module that extracts joint-level information and a second frame-level module capturing temporal information via convolutional layers and spatial and temporal max-pooling. Our classification branch borrows ideas from Mao \textit{et al.}'s~\cite{Mao19} and Zhang \textit{et al.}'s~\cite{zhang20e} architectures. It is composed of graph convolutional blocks as proposed by Mao \textit{et al.}~\cite{Mao19} combined with the frame-level module architecture proposed by Zhang \textit{et al.}~\cite{zhang20e}. 
\subsection{Physical Exercise Analysis}
Physical exercise analysis aims to prevent injuries that may arise when a person performs motions incorrectly. In its simplest form, such an analysis amounts to detecting whether the subject performs the exercise correctly or not. This was achieved several works~\cite{Chen20d,Rangari22,Dittakavi22} by exploiting 2D poses extracted from the input images. In particular, Dittakavi \textit{et al.}~\cite{Dittakavi22} detected which joints need to be fixed by finding the overall joint angle distribution of the dataset and detecting poses in which a joint angle is an anomaly. This framework operates on single frames, as opposed to our method which operates on entire sequences. In~\cite{Zell17}, Zell \textit{et al.} represented the human body as a mass-spring model and analyzed the extension torque on certain joints, allowing them to classify whether a motion is performed correctly or not. While useful, such classification-based approaches offer limited information to the user, as they do not provide them with any feedback about the specific type of mistakes they made. Moreover, most of existing works operate on 2D pose inputs~\cite{Chen20d,Yang21,Kanase21,Rangari22,Dittakavi22}. Similar to ~\cite{Fieraru21}, we also design our framework to work with 3D poses enabling us to be robust to ambiguities found in 2D poses.
While a few works took some steps toward giving feedback~\cite{Yang21,Kanase21,Fieraru21}, this was achieved in a hard-coded fashion, by thresholding angles between some of the body joints. As such, this approach relies on manually defining such thresholds, and thus does not easily extend to new exercises. Furthermore, it does not provide the user personalized corrective measures in a visual manner, by demonstrating the correct version of their performance. We address this by following a data driven approach able to automatically learn the different ``correct" forms of an exercise, and that can easily extend to different types of exercises and mistakes. To the best of our knowledge, our framework is the first to both identify mistakes and suggest personalized corrections to the user.
\section{Methodology}

Before we introduce our framework in detail, let us formally define the tasks of motion classification and correction. Motion classification seeks to predict the action class $c$ of a sequence of 3D poses from $t=1$ to $t=N$, denoted as $\mathbf{P}_{1:N}$. We can write this as
\begin{align*}
c = F_\text{class}(\mathbf{P}_{1:N})\;,
\end{align*}
where $F_\text{class}$ is the classification function.

We define motion correction as the task of finding the ``correct" version of a sequence, which can be written as
\begin{align*}
\hat{\mathbf{P}}_{1:N} = F_\text{corr}(\mathbf{P}_{1:N})\;,
\end{align*}
where $F_\text{corr}$ is the correction function and $\hat{\mathbf{P}}_{1:N}$ is the corrected sequence. Ideally, the corrected sequence should be of class ``correct", i.e.
\begin{align*}
c_\text{correct} = F_\text{class}(\hat{\mathbf{P}}_{1:N})\;,
\end{align*}
where $c_\text{correct}$ is the label corresponding to a correct action.

Given these definitions, we now describe the framework we designed to address these tasks and discuss our training and implementation details. 

\subsection{Exercise Analysis Framework}

Our framework for providing exercise feedback relies on GCNs and consists of two branches: One that predicts whether the input motion is correct or incorrect, specifying the mistake being made in the latter case, and one that outputs a corrected 3D pose sequence, providing a detailed feedack to the user. We refer to these two branches as the ``classifier" and ``corrector" models, respectively.

Inspired by Mao \textit{et al.}~\cite{Mao19}, we use the DCT coefficients of joint trajectories, rather than the 3D joint positions, as input to our model. This allows us to easily process sequences of different lengths.

The corrector model outputs DCT coefficient residuals, which are then summed with the input coefficients and undergo an inverse-DCT transform to be converted back to a series of 3D poses.

\begin{figure}[!]
	\centering
	\includegraphics[width=1.0\linewidth]{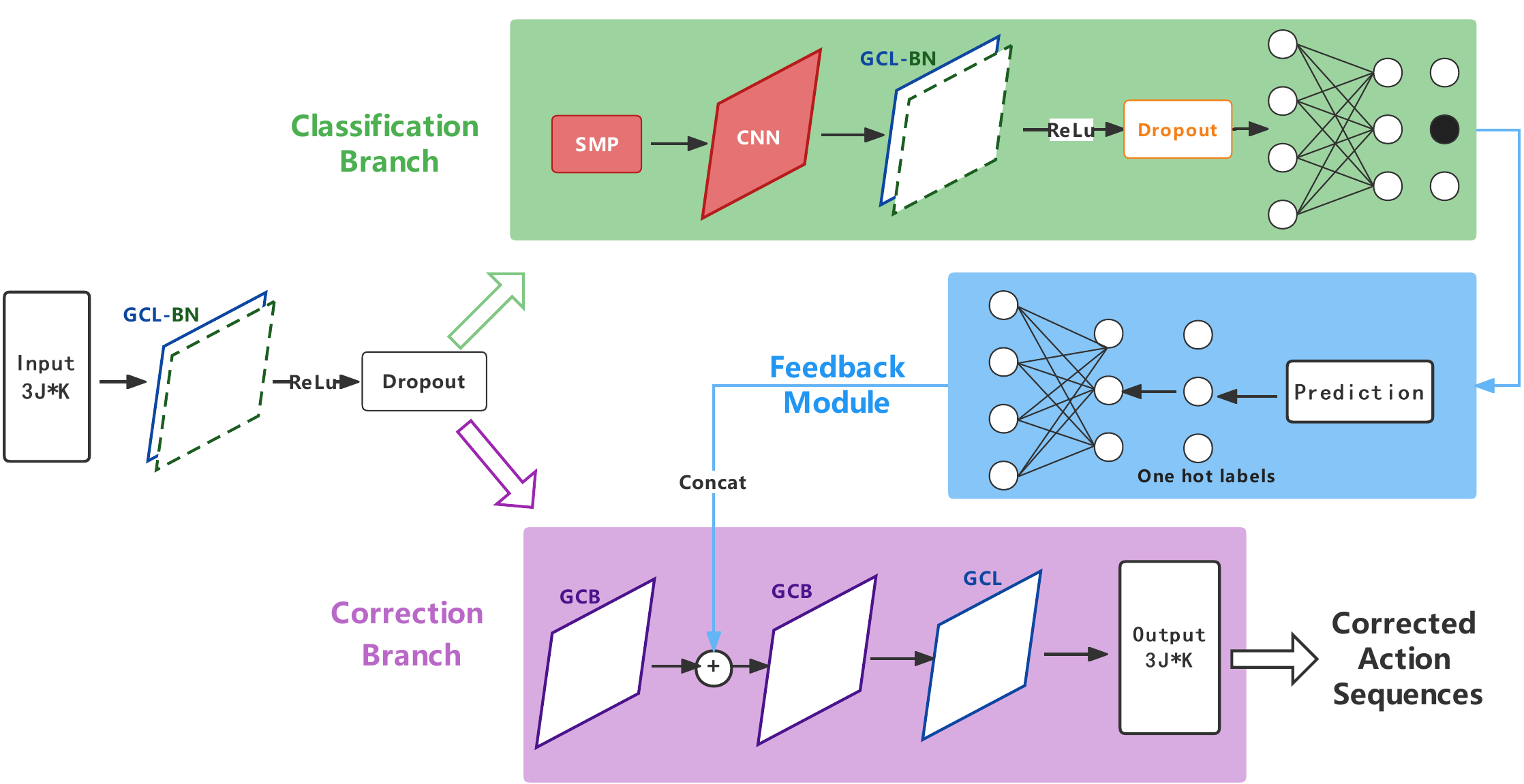}
	\caption{{\bf Our framework} consists of a classification and a correction branch. They share several graph convolutional layers are then split such that the classification branch identifies the type of mistakes made by the user and the correction branch outputs a corrected pose sequence. The result of the classification branch is fed to the correction branch via a feedback module.} 
	\label{fig:overall}
\end{figure}

To reduce the time and space complexity of training the classifier and the corrector separately and to improve the accuracy of the model, we combine the classification and correction branches into a single end-to-end trainable model. Figure~\ref{fig:overall} depicts our overall framework. It takes the DCT coefficients of each joint trajectory as input. The first layers are shared by the two models, and the framework then splits into the classification and correction branches.

Furthermore, we feed the predicted action labels coming from the classification branch to the correction branch. 
We depict this in Figure~\ref{fig:overall} as the ``Feedback Module". Specifically, we first find the label with the maximum score predicted by the classification branch,  convert this label into a one-hot encoding, and feed it to a fully-connected layer. The resulting tensor is concatenated to the output of the first graph convolutional blocks (GCB) of the correction branch. This process allows us to explicitly provide label information to the correction module, enabling us to further improve the accuracy of the corrected motion.

\subsubsection{Implementation and Training Details.}

\begin{figure}[!]
	\centering
	\includegraphics[width=1.0\linewidth]{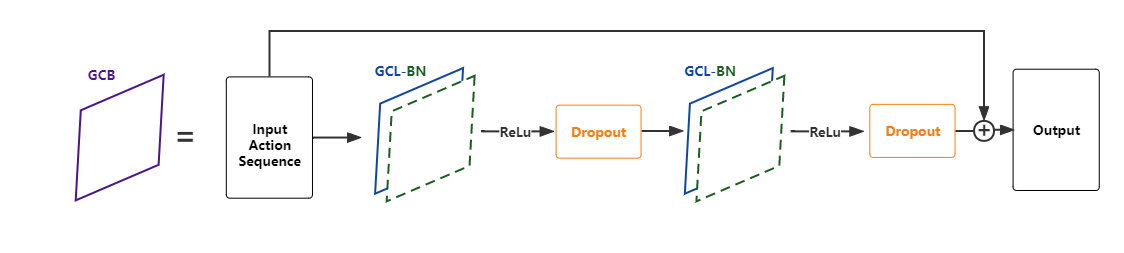}
	\caption{{\bf Graph Convolutional Block (GCB)} consisting of graph convolutional layers, batch normalization layers, ReLUs and drop-outs.} 
	\label{fig:gcb}
\end{figure}

We primarily use GCB similar to those presented in~\cite{Mao19} in our network architecture, depicted in Figure~\ref{fig:gcb}. These modules allow us to learn the connectivity between different joint trajectories. Each graph convolutional layer is set to have $256$ hidden features. Additionally, our classification branch borrows ideas from Zhang \textit{et al.}'s~\cite{zhang20e} action recognition model. It is a combination of GCB modules and the frame-level module architecture of~\cite{zhang20e} consisting of convolutional layers and spatial max-pooling layers. 

We train our network in a supervised manner by using pairs of incorrectly performed and correctly performed actions. However, it is not straightforward to find these pairs of motions. The motion sequences are often of different lengths, and we face the task of matching incorrectly performed actions to the closest correctly performed action from the same actor. To do so, we make use of Dynamic Time Warping (DTW)~\cite{Sakoe78}, which enables us to find the minimal alignment cost between two time series of different lengths, using dynamic programming. We compute the DTW loss between each incorrect and correct action pair candidate and select the pair with the smallest loss value. 
We use the following loss functions to train our model. 
\begin{itemize}
	\item $E_\text{corr}$: The loss of the correction branch, which aims to minimize the soft-DTW~\cite{Cuturi17} loss between the corrected output sequence and the closest correct motion sequence, determined as described previously. The soft-DTW loss is a differentiable version of the DTW loss, implemented by replacing the minimum operation by a soft minimum.
	\item $E_\text{smooth}$: The smoothness loss on the output of the correction branch, to ensure the produced motion is smooth and realistic. It penalizes the velocities of the output motion by imposing an L2 loss on them.
	\item $E_\text{class}$: The loss of the classification branch, which aims to minimize the cross entropy loss between the predicted logits and the ground-truth instruction label.
\end{itemize}

We combine these losses into
\begin{equation}
E_\text{loss} = w_\text{corr} E_\text{corr} + w_\text{class} E_\text{class} +  w_\text{smooth} E_\text{smooth} ,
\end{equation}
where $E_\text{loss}$ is the overall loss and $w_\text{corr}, w_\text{class}, w_\text{smooth}$ are the weights of the correction, classification, and smoothness losses, respectively. For our experiments we set $w_\text{corr}=1$, $w_\text{class}=1$, and $w_\text{smooth}=1e-3$.

During training, we use curriculum learning in the feedback module: Initially the ground-truth instruction labels are given to the correction branch. We then use a scheduled sampling strategy similar to~\cite{Bengio15b}, where the probability of using the ground-truth labels instead of the predicted ones decreases from 1 to 0 linearly as the epochs increase. In other words, the ground-truth labels are progressively substituted with the labels predicted by the classification branch, until only the predicted labels are used. During inference, only the predicted labels are given to the correction branch.

We use Adam~\cite{Kingma15} as our optimizer. The learning rate is initially set to $0.01$ and decays according to the equation $\text{lr} = 0.01 \cdot 0.9^{i/s}$, where $\text{lr}$ is the learning rate, $i$ is the epoch and $s$ is the decay step, which is set to 5. To increase robustness and avoid overfitting, we also use drop-out layers with probability $0.5$. We use a batch size of $32$ and train for $50$ epochs.

\section{EC3D Dataset}

To evaluate our approach, we recorded and processed a dataset of physical exercises performed both correctly and incorrectly, and named the ``EC3D" (\textbf{E}xercise \textbf{C}orrection in \textbf{3D}) dataset. 

Specifically, this dataset contains $3$ types of actions, each with $4$ subjects who repeatedly performed a particular correct or incorrect motion as instructed. We show the number of sequences per action and the instructions for each subject in Table~\ref{tab:dataset}. The dataset contains a total of $132$ squat, $127$ lunge, and $103$ plank action sequences, split across $11$ instruction labels.

\begin{figure}[h!]
	\centering
	\includegraphics[width=1.0\linewidth]{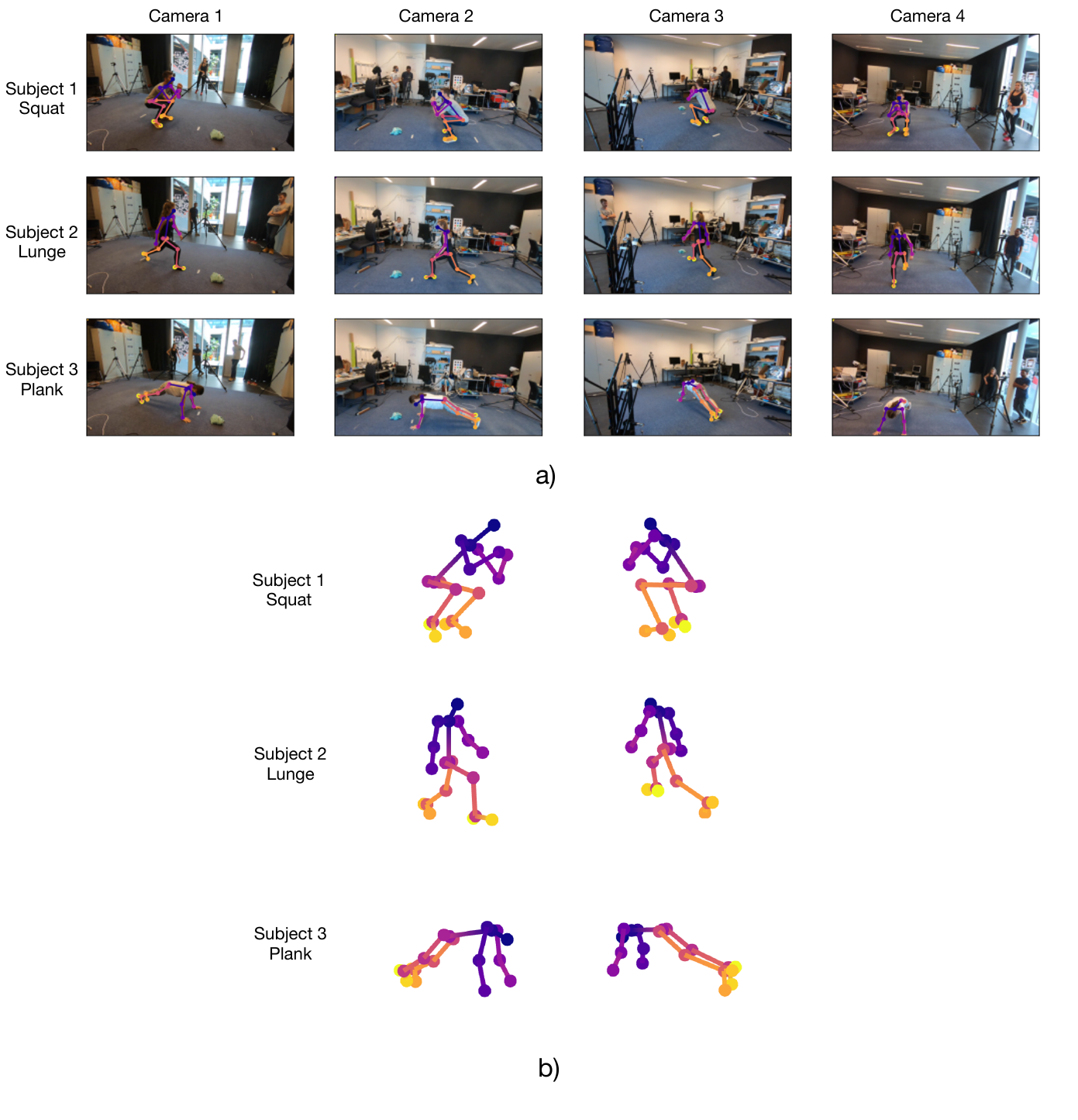}
	\caption{{\bf Examples images from the EC3D dataset,} depicting images from the SQUAT, lunge, and plank classes with their corresponding 3D pose visualizations. a) Images for each exercise type from the dataset from each camera viewpoint, with the 2D poses overlayed. b) The corresponding 3D poses, visualized from two different viewpoints.}
	\label{fig:dataset}
\end{figure}

The videos were captured by $4$ GoPro cameras placed in a ring around the subject, using a frame rate of $30$ fps and a $1920 \times 1080$ image resolution. Figure~\ref{fig:dataset} depicts example images taken from the dataset with their corresponding 2D and 3D skeleton representation. The cameras' intrinsics were obtained by recording a chessboard pattern and using standard calibration methods implemented in OpenCV~\cite{OpenCV}.

\begin{table}[]
\begin{tabular}{|l|l|l|l|l|l|l|l|}
\hline
Exercise                & \begin{tabular}[c]{@{}l@{}}Instruction\\ Label\end{tabular} & Subject 1 & Subject 2 & Subject 3 & Subject 4 & \begin{tabular}[c]{@{}l@{}}Total \\ (per instruction)\end{tabular} & \begin{tabular}[c]{@{}l@{}}Total \\ (per action)\end{tabular} \\ \hline
\multirow{5}{*}{Squats} & Correct                                                     & 10        & 10        & 11        & 10        & 41                                                                 & \multirow{5}{*}{132}                                          \\ \cline{2-7}
                        & Feet too wide                                               & 5         & 8         & 5         & 5         & 23                                                                 &                                                               \\ \cline{2-7}
                        & Knees inward                                                & 6         & 7         & 5         & 5         & 23                                                                 &                                                               \\ \cline{2-7}
                        & Not low enough                                              & 5         & 7         & 5         & 4         & 21                                                                 &                                                               \\ \cline{2-7}
                        & Front bent                                                  & 5         & 6         & 6         & 7         & 24                                                                 &                                                               \\ \hline
\multirow{3}{*}{Lunges} & Correct                                                     & 12        & 11        & 11        & 12        & 46                                                                 & \multirow{3}{*}{127}                                          \\ \cline{2-7}
                        & Not low enough                                              & 10        & 10        & 10        & 10        & 40                                                                 &                                                               \\ \cline{2-7}
                        & Knee passes toe                                             & 10        & 10        & 11        & 10        & 41                                                                 &                                                               \\ \hline
\multirow{3}{*}{Planks} & Correct                                                     & 7         & 8         & 11        & 7         & 33                                                                 & \multirow{3}{*}{103}                                          \\ \cline{2-7}
                        & Arched back                                                 & 5         & 5         & 11        & 9         & 30                                                                 &                                                               \\ \cline{2-7}
                        & Hunch back                                                  & 10        & 10        & 11        & 9         & 40                                                                 &                                                               \\ \hline
\end{tabular}

\caption{\textbf{The EC3D dataset} with the number of sequences per instruction of each subject, the total number of sequences per instruction and the total number of sequences per action. We reserve Subjects 1, 2, and 3 for training and 4 for testing.}
\label{tab:dataset}

\end{table}

We annotated the 3D poses in an automated manner, whereas the action and instruction labels were annotated manually. Specifically, the 3D pose annotation was performed as follows: First, the 2D joint positions were extracted from the images captured by each camera using OpenPose~\cite{Cao17}, an off-the-shelf 2D pose estimation network. We then used bundle adjustment to determine the cameras' extrinsics. For the bundle adjustment algorithm to converge quickly and successfully, additional annotations were made on static landmarks in $5$ frames. Since the cameras were static during recording, for each camera, we averaged the extrinsics optimized for each of these frames. Afterwards, these values were kept constant, and we triangulated the 2D poses to compute the 3D poses. 

During the triangulation process, we detected whenever any joint had a high reprojection error to catch mistakes in the 2D pose estimates. Such 2D pose annotations were discarded to prevent mistakes in the 3D pose optimization. The obtained 3D pose values were afterwards smoothed using a Hamming filter to avoid jittery motion. Finally, we manually went through the extracted 3D pose sequences in order to ensure that there are no mistakes and that they are consistent with the desired motion.

To make the resulting 3D poses uniform, we further normalized, centred and rotated them. As the different heights and body sizes of the different subjects cause differences in skeletal lengths, a random benchmark was selected to normalize the skeletal lengths while maintaining the connections between joints. Furthermore, we centered all poses on their hip joint and rotated them so that the spine was perpendicular to the ground and all movements performed in the same direction.

\section{Evaluation}

\subsection{Dataset and Metrics}
We use the EC3D dataset to evaluate our model performance both quantitatively and qualitatively. We use subjects 1, 2, and 3 for training and subject 4 for evaluation.

We use top-1 classification accuracy to evaluate the results of the instruction classification task, as used by other action classification works~\cite{Li19m,zhang20e}. For the motion correction task, we make use of the action classifier branch: If the corrected motion is classified as ``correct" by our classification branch, we count the correction as successful. We report the percentage of successfully corrected motions as the evaluation metric for this task.

\subsection{Quantitative Results}
We achieve an average mistake recognition accuracy of $90.9\%$ when classifying sequences in EC3D, as shown by the detailed results for each specific exercise instruction in Table~\ref{tab:results_class}. In the same table, we also show that $94.2\%$ of the corrected results are classified as ``correct" by our classification model. The high classification accuracy and correction success show that our framework is indeed capable of analyzing physical exercises and giving useful feedback.

\begin{table}[]
\centering
\begin{tabular}{|ll|l|l|}
\hline
\multicolumn{1}{|l|}{Exercise}                & Mistake Label   & \begin{tabular}[c]{@{}l@{}}Classification\\ Accuracy (\%)\end{tabular} & \begin{tabular}[c]{@{}l@{}}Correction\\ Success (\%)\end{tabular} \\ \hline
\multicolumn{1}{|l|}{\multirow{5}{*}{Squats}} & Correct         & 90.0                                                                   & 100                                                               \\ \cline{2-4} 
\multicolumn{1}{|l|}{}                        & Feet too wide   & 100                                                                    & 100                                                               \\ \cline{2-4} 
\multicolumn{1}{|l|}{}                        & Knees inward    & 100                                                                    & 100                                                               \\ \cline{2-4} 
\multicolumn{1}{|l|}{}                        & Not low enough  & 100                                                                    & 100                                                               \\ \cline{2-4} 
\multicolumn{1}{|l|}{}                        & Front bent      & 57.1                                                                   & 85.7                                                              \\ \hline
\multicolumn{1}{|l|}{\multirow{3}{*}{Lunges}} & Correct         & 66.7                                                                   & 100                                                               \\ \cline{2-4} 
\multicolumn{1}{|l|}{}                        & Not low enough  & 100                                                                    & 60.0                                                              \\ \cline{2-4} 
\multicolumn{1}{|l|}{}                        & Knee passes toe & 100                                                                    & 90.0                                                              \\ \hline
\multicolumn{1}{|l|}{\multirow{3}{*}{Planks}} & Correct         & 85.7                                                                   & 100                                                               \\ \cline{2-4} 
\multicolumn{1}{|l|}{}                        & Arched back     & 100                                                                    & 100                                                               \\ \cline{2-4} 
\multicolumn{1}{|l|}{}                        & Hunch back      & 100                                                                    & 100                                                               \\ \hline
\multicolumn{2}{|l|}{Average}                                   & 90.9                                                                   & 94.2                                                              \\ \hline
\end{tabular}
\caption{\textbf{Results of our classification and correction branches on the EC3D dataset.} We achieve $90.9\%$ recognition accuracy on average and successfully correct $94.2\%$ of the mistakes.}
\label{tab:results_class}

\end{table}

As no existing works have proposed detailed correction strategies, we compare our framework to a simple correction baseline consisting of retrieving the closest ``correct" sequence from the training data. The closest sequence is determined as the sequence with the lowest DTW loss value to the input sequence. In Table~\ref{tab:results_corr}, we provide the DTW values between the incorrectly performed input and the corrected output. For this metric, the lower, the better, i.e., the output motion should be as close as possible to the original one while being corrected as necessary. Our framework yields a high success rate of correction together with a lower DTW loss than the baseline, thus supporting our claims. Note that we do not evaluate the baseline's correction success percentage because it retrieves the same sequences that were used to train the network, to which the classification branch might have already overfit.

\begin{table}[]
\centering

\begin{tabular}{|ll|l|l|}
\hline
\multicolumn{1}{|l|}{Exercise}                & Mistake Label   & Retrieval Baseline & Our Framework \\ \hline
\multicolumn{1}{|l|}{\multirow{5}{*}{Squats}} & Correct         & 1.28               & \textbf{0.56} \\ \cline{2-4} 
\multicolumn{1}{|l|}{}                        & Feet too wide   & 4.23               & \textbf{1.46} \\ \cline{2-4} 
\multicolumn{1}{|l|}{}                        & Knees inward    & 1.61               & \textbf{0.66} \\ \cline{2-4} 
\multicolumn{1}{|l|}{}                        & Not low enough  & 1.83               & \textbf{0.61} \\ \cline{2-4} 
\multicolumn{1}{|l|}{}                        & Front bent      & 4.74               & \textbf{2.53} \\ \hline
\multicolumn{1}{|l|}{\multirow{3}{*}{Lunges}} & Correct         & 1.94               & \textbf{1.82} \\ \cline{2-4} 
\multicolumn{1}{|l|}{}                        & Not low enough  & 1.86               & \textbf{1.31} \\ \cline{2-4} 
\multicolumn{1}{|l|}{}                        & Knee passes toe & 2.27               & \textbf{1.48} \\ \hline
\multicolumn{1}{|l|}{\multirow{3}{*}{Planks}} & Correct         & 2.41               & \textbf{1.79} \\ \cline{2-4} 
\multicolumn{1}{|l|}{}                        & Arched back     & 12.20              & \textbf{1.53} \\ \cline{2-4} 
\multicolumn{1}{|l|}{}                        & Hunch back      & 4.10               & \textbf{1.09} \\ \hline
\multicolumn{2}{|l|}{Average}                                   & 3.49               & \textbf{1.35} \\ \hline
\end{tabular}

\caption{\textbf{DTW results of the correction branch.} We compare our framework to a simple baseline retrieving the best matching ``correct" sequence from the training dataset depending on the classification label. We report the DTW loss between the input and the output sequences (lower is better). Our framework successfully corrects the subject's mistakes, while not changing the input so drastically that the subject would not be able to recognize their own performance.}
\label{tab:results_corr}

\end{table}

\subsection{Qualitative Results}

\begin{figure}[]
	\centering
	\vspace{3em}
	\includegraphics[width=1.0\linewidth]{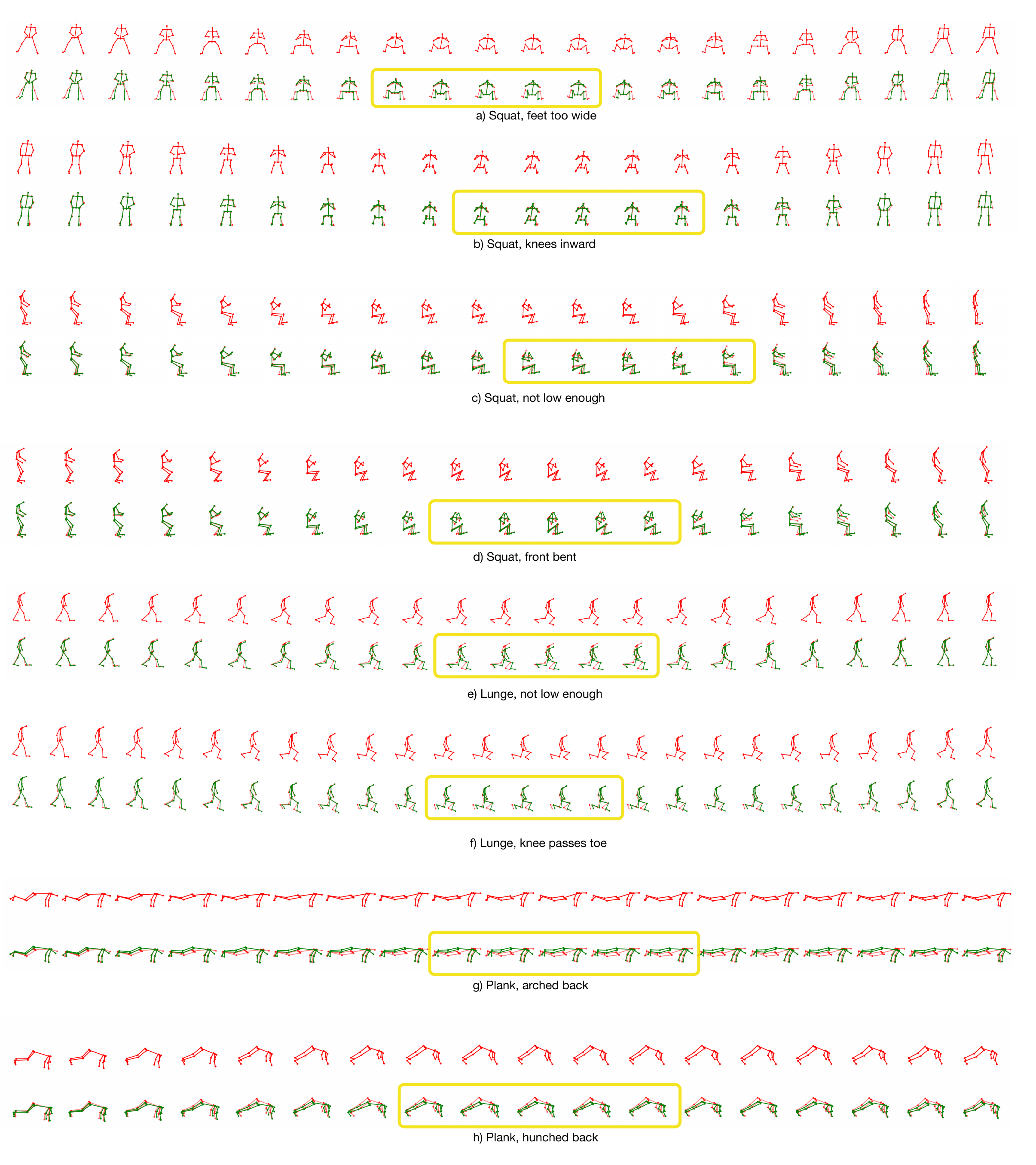}
	\caption{{\bf Qualitative results from our framework} depicting incorrect input motions and corrected output motions from categories a-d) squats, e-f) lunges, g-h) planks. We present the incorrect input sequences (red) in the top row. The corrected sequences (green) overlaid on top of the incorrect input sequences (red) are presented in the bottom row. The most significant corrections are highlighted with a yellow bounding box. We find that our proposals are successful in correcting the incorrect sequences. This figure is best viewed in color and zoomed in on a screen.}
    \vspace{4em}

	\label{fig:qualitative}
\end{figure}

In Figure~\ref{fig:qualitative}, we provide qualitative results corresponding to all the incorrect motion examples from each action category. Note that the incorrect motions are successfully corrected, yet still close to the original sequence. This makes it possible for the user to easily recognize their own motion and mistakes.

\subsection{Ablation Studies}

\begin{figure}[!]
	\centering
	\includegraphics[width=1.0\linewidth]{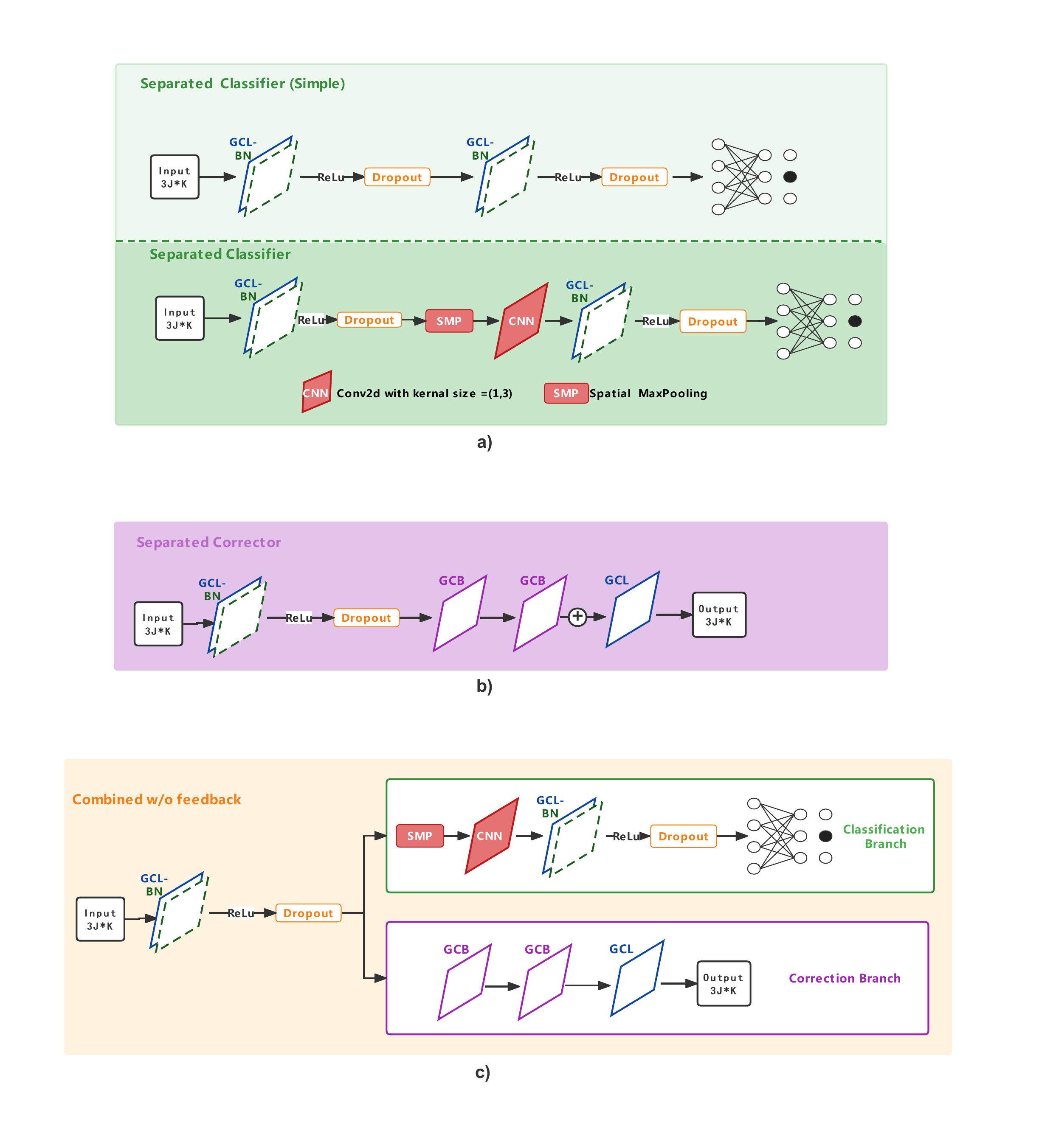}
	\caption{{\bf Ablation study frameworks.} We depict the different architectures we evaluated for the ablation studies: a) Separated classifier (simple) and separated classifier. b) Separated corrector. c) Combined model without feedback. This model does not include a ``Feedback Module", the classification branch's results are not explicitly fed to the correction branch.} 
	\vspace{3em}

	\label{fig:ablation}
\end{figure}

 We have tried various versions of our framework and recorded our results in Table~\ref{tab:results_ablation}. In this section, we present the different experiments, also depicted in Figure~\ref{fig:ablation}, and the discussions around these experiments.

 \textbf{Separated models.} We first analyze the results of separated classification networks. According to Table~\ref{tab:results_ablation}, our separated classification branch architecture is denoted as ``separated classification." We have also evaluated a simpler, fully GCN based separated action classifier branch, denoted as ``separated classification (simple)". We show that the results of the classification branch degrade slightly when separated from the correction branch. This indicates that the classification branch also sees a minor benefit from being part of a combined model. The simpler classification network performs worse than our architecture inspired by~\cite{zhang20e}, showing that the pooling module improves the classification accuracy. 

Afterwards, we analyze the results of a separated correction network, denoted as ``separated correction". Here the difference is quite profound; we see that separating the correction model from the classification model degrades correction success significantly. We note that $50$ epochs was not enough for the separated corrector framework to converge, therefore we trained it for a total of $150$ epochs.

\textbf{Combined models.} 

We evaluate our framework without the feedback module (``combined w/o feedback"), and train without the smoothness loss (``combined w/o smoothness"). We find that these perform worse than our model with the feedback module and with the smoothness loss in terms of correction success. This shows that using the classification results as feedback as well as the smoothness loss for training is useful for more successful corrections. We notice that our framework trained without smoothness loss has higher classification accuracy, despite having a lower correction success. We believe this is due to the fact that the smoothness loss acts as a regularizer on the framework, therefore causing slight performance losses to the classification branch. However, the results of the correction success are significantly higher with the smoothness loss. 

Further qualitative and quantitative results are presented in the supplementary material.

\begin{table}[]
\centering
\begin{tabular}{|l|l|l|l|}
\hline
                           &                                      & \begin{tabular}[c]{@{}l@{}}Classification\\ Accuracy (\%)\end{tabular} & \begin{tabular}[c]{@{}l@{}}Correction\\ Success (\%)\end{tabular} \\ \hline
\multirow{4}{*}{Separated} & Separated classification (simple)    & 88.6   & -         \\ \cline{2-4} 
                           & Separated classification             & 89.8   & -       \\ \cline{2-4} 
                           & Separated correction                 & -      & 83.5   \\ \hline
\multirow{3}{*}{Combined}  
                           & Combined w/o feedback                & 82.3    & 85.3   \\ \cline{2-4} 
                           & Combined w/o smoothness          & \textbf{93.4}   & 87.5  \\ \cline{2-4} 
                           & Ours                                 & 90.9    & \textbf{94.2}     \\ \hline
\end{tabular}

\caption{\textbf{Results of the ablation studies} with several variations of our framework. We report the classification accuracy (\%) and the correction success rate (\%), where higher is better for both metrics. Our framework benefits greatly from combining the two tasks in a end-to-end learning fashion, from using a feedback module, and from using a pooling layer in the classification branch. The smoothness loss causes slight degradation in classification accuracy but is greatly beneficial for the correction success.}
\label{tab:results_ablation}

\end{table}

\subsection{Limitations and Future Work}

While our current results are quite impressive, there is still room for improvement in terms of performance. In particular, our framework struggles in correcting specific types of motions, such as not low enough lunges. We plan to explore different additions to our framework, such as an attention module, to better correct these types of mistakes. 

Our future work will consist of expanding the dataset to include more action sequences and more types of mistakes, performed by a larger set of subjects. We believe that this will allow us to further improve our framework and add components that address the shortcomings we will discover using such a dataset.

\section{Conclusion}

We have presented a 3D pose based feedback framework for physical exercises. We have designed this framework to output feedback in two branches; a classification branch to identify a potential mistake and a correction branch to output a corrected sequence. Through ablation studies, we have validated our network architectural choices and presented detailed experimental results, making a strong case for the soundness of our framework design.
We have also introduced a dataset of physical exercises, on which we have achieved $90.9\%$ classification accuracy and $94.2\%$ correction success.

{\small
	\bibliographystyle{splncs}
	\bibliography{string,vision,graphics,learning,robotics,misc}
}

\section{Supplementary Material}

\subsection{Videos}

Our video accessible at \url{https://youtu.be/W3kyyeHe0SI} gives an overview of our paper by introducing our problem, explaining our methodology and presenting our results.

\subsection{Further Ablation Studies}
\textbf{Ablation study on TMP module.} The architecture of our classification branch is highly inspired by the frame-level module architecture proposed by Zhang \textit{et al.} (SGN)~\cite{zhang20e}. The set-up used by SGN is a spatial MaxPooling (SMP) layer, followed by two convolutional layers and a temporal MaxPooling layer (TMP). Our classification branch omits the TMP layer, because we have found it to not be useful for our case. We suspect that the reason for this is we do not have temporal data as input to our networks, instead we have DCT coefficients. In Table~\ref{tab:ablation_TMP} we present an ablation study on using a TMP layer. The results obtained using a TMP layer are worse in terms of both classification accuracy and correction success.

\begin{table}[]
\centering{
\begin{tabular}{|l|l|l|l|l|}
\hline
               & \begin{tabular}[c]{@{}l@{}}Classification\\ Accuracy(\%)\end{tabular} & \begin{tabular}[c]{@{}l@{}}Classification\\ Loss\end{tabular} & \begin{tabular}[c]{@{}l@{}}Correction\\ Accuracy(\%)\end{tabular} & \begin{tabular}[c]{@{}l@{}}Correction\\ Loss\end{tabular} \\ \hline
Ours with TMP  & 76.1                                                                  & 1.62                                                          & 91.5                                                              & 3.35                                                      \\ \hline
Ours (w/o TMP) & \textbf{90.9}                                                         & 1.30                                                          & \textbf{94.2}                                                     & 1.45                                                      \\ \hline
\end{tabular}
}
\caption{\textbf{Ablation study on the TMP layer.} We show the results of our framework trained using a TMP module and without using a TMP module (ours). We note that the TMP module only plays a role in the classification branch, however this still affects the performance of correction also, as the two branches are trained as part of a single network. We find that the results are worse for both classification accuracy and correction success when the TMP module is included.}
\label{tab:ablation_TMP}
\end{table}

\textbf{Ablation study on the smoothness loss.} We have trained our network with different weights for the smoothness loss term, denoted as $w_\text{smooth}$. The results are reported in Table ~\ref{tab:ablation_smoothness}. We find that $w_\text{smooth}=1e-3$ gives the best results in terms of the correction success. However the results obtained when $w_\text{smooth}=0$ gives slightly better results for classification accuracy.

\begin{table}[]
\centering
\begin{tabular}{|l|l|l|}
\hline
                & \begin{tabular}[c]{@{}l@{}}Classification \\ Accuracy(\%)\end{tabular}  & \begin{tabular}[c]{@{}l@{}}Correction\\ Success(\%)\end{tabular}\\ \hline
 $w_\text{smooth}=1e-1$        & 81.8              & 72.7       \\ \hline
 $w_\text{smooth}=1e-2$       & 81.8               & 85.9       \\ \hline
$w_\text{smooth}=1e-3$ \textbf{(Ours)}    & 90.9    & \textbf{94.2}             \\ \hline
 $w_\text{smooth}=1e-5$      & 81.8         & 85.0       \\ \hline
 $w_\text{smooth}=0$          & \textbf{93.4}   & 87.5  \\ \hline

\end{tabular}
\caption{\textbf{Ablation study on the smoothness loss.} We present the results of our network when trained with different weights for the smoothness loss. We find that setting the weight of the smoothness loss to $1e-3$ gives the best results in terms correction success. The classification accuracy is higher when smoothness loss is not used.}
\label{tab:ablation_smoothness}
\end{table}

\subsection{Classification on the NTU RGB+D Dataset}

\textbf{NTU RGB+D Dataset.} We use the NTU RGB+D~\cite{shahroudy16}, a widely-used dataset for evaluating the performance of action classification networks~\cite{shahroudy16,Li19m,zhang20e}. We use cross-subject division to split the training set and test set according to the person ID. A total of $40$ subjects were divided into a training set of $17$ subjects, a validation set of $3$ subjects and a test set of $20$ subjects. 
 
NTU RGB+D contains $56,880$ action samples. As stated in the official dataset release, 302 samples in the dataset dataset have missing or incomplete skeleton data. In addition, some of the actions involve two people interacting with each other, which do not match the expected input to the model, and the video frame lengths are not uniform across sequences. To overcome these challenges, we have pre-processed this dataset.
In order to remove noisy data, for the missing or incomplete skeleton data action sequences, we used the official list of missing data indices. As they only represent $0.53\%$ of the total data volume, these lossy data are directly removed from the dataset.

For some of the sequences, there are two-subjects in a single frame. For such sequences we consider the pose coordinates of only one of them. Since the two people motions involve movements which are mirror-symmetrical (e.g. A55 Hugging, A59 Walking towards, etc...), we find this method to be sufficient.

Different motion sequences can have different lengths. In order to counter different video frame rates, SGN tries to segment the entire skeleton sequence into 20 clips equally, and randomly select one frame from each clip to have a new sequence of 20 frames. However, in order to maintain consistency between the NTU dataset and the EC3D data, we modify the inputs when feeding our data to our model by finding their top $25$ DCT coefficients. By doing so, for all input sequences we have $25$ DCT coeffiecients and do not have to worry about the differences in sequence lengths. We then normalize, centralize and rotate the dataset as we have done with EC3D.

\textbf{Classification results on NTU RGB+D Dataset.} We present our classification branch's results on the NTU RGB+D dataset in Table~\ref{tab:results_class_ntu} to compare our model's performance to SOTA action classification methods. Although we do not achieve SOTA performance, our performance is comparable to the results of many of these methods, showing that it is a reliable, lightweight method for action classification on other mainstream datasets. We note that our goal is not to achieve SOTA action classification, but to achieve high enough accuracy to give reliable feedback and to also evaluate the results of the action correction branch.

\begin{table}[]
\centering
\begin{tabular}{|l|l|}
\hline
Methods         & \begin{tabular}[c]{@{}l@{}}Classification \\ Accuracy (\%)\end{tabular} \\ \hline
HBRNN-L~\cite{Du15}         & 59.1                                                                    \\ \hline
Part-Aware LSTM~\cite{shahroudy16} & 62.9                                                                    \\ \hline
ST-LSTM + Trust Gate~\cite{Liu16c} & 69.2                                                                 \\ \hline
STA-LSTM~\cite{Song17}        & 73.4                                                                    \\ \hline
GCA-LSTM~\cite{Liu17}        & 74.4                                                                    \\ \hline
DPRL+GCNN~\cite{Tang18e}       & 83.5                                                                    \\ \hline
HCN~\cite{Li19m}             & 86.5                                                                    \\ \hline
AS-GCN~\cite{Li19n} & 86.8                                                                    \\ \hline
VA-CNN~\cite{Zhang19f}  & 88.7                                                                    \\ \hline
SGN~\cite{zhang20e}             & \textbf{89.0}                                                                    \\ \hline
Ours            & 70.1                                                                    \\ \hline
\end{tabular}
\caption{\textbf{The classification branch's results on the NTU RGB+D dataset.} While we do not achieve SOTA performance, we outperform HBRNN-L~\cite{Du15}, Part-Aware LSTM~\cite{shahroudy16}, and ST-LSTM + Trust Gate~\cite{Liu16c} . Our results are comparable to those of STA-LSTM~\cite{Song17}  and GCA-LSTM~\cite{Liu17}. We conclude that our network is able to achieve acceptable performance on larger, more mainstream datasets as well.}
\label{tab:results_class_ntu}
\end{table}

\subsection{Further Qualitative Results}
\begin{figure}[]
	\centering
	\vspace{3em}
	\includegraphics[width=1.0\linewidth]{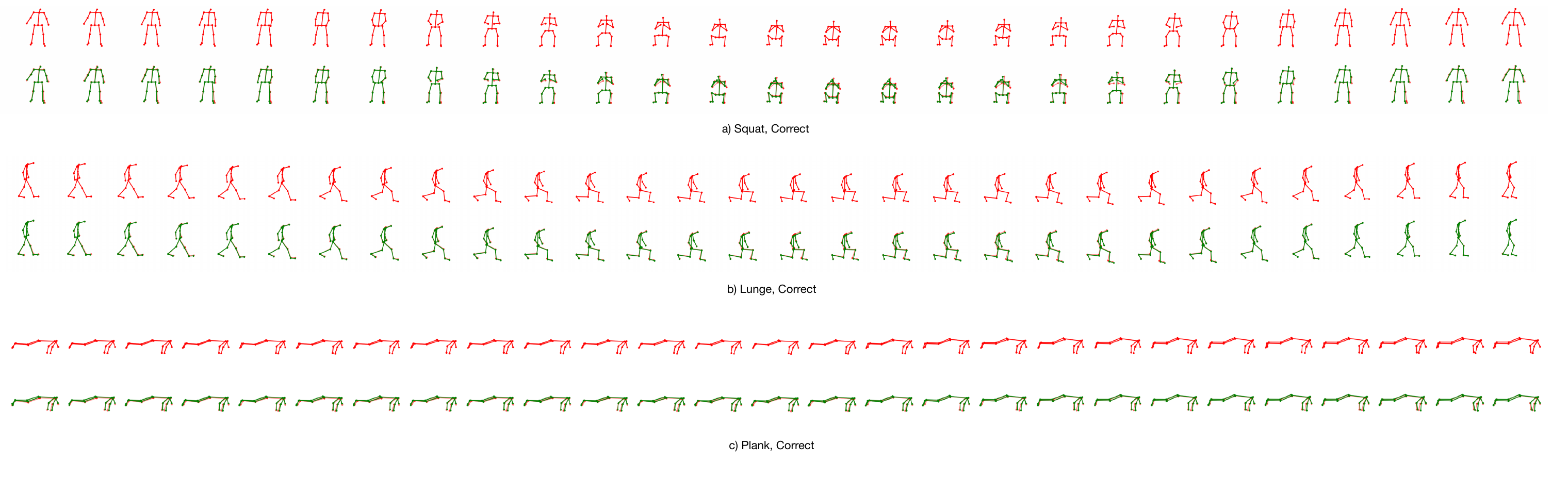}
	\caption{{\bf Qualitative results from our framework of correcting already ``correct" motions.} We input motions and corrected output motions from categories a) squats, b) lunges, c) planks. We present the input sequences (red) in the top row. The corrected sequences (green) overlaid on top of the input sequences (red) are presented in the bottom row. We find that since the input sequences are already correct, the adjustments made by our framework are very minor. This figure is best viewed in color and zoomed in on a screen.}
    \vspace{4em}

	\label{fig:qualitative_correct}
\end{figure}

We have evaluated the behaviour of our framework on inputs that are already correct and we present qualitative results in Figure~\ref{fig:qualitative_correct}. We find that since the input sequences are already correct, the framework's adjustments are very minor. 

\subsection{Detailed Quantitative Results}
We present detailed quantitative results of our model in Tables~\ref{tab:classification_results} and ~\ref{tab:correction_result}. Table~\ref{tab:classification_results} presents a confusion matrix of the results of our classification branch. We can clearly see which actions are confused with which ones. We find that the ``correct squat" is once confused with a ``front bent squat" and the ``correct plank" is confused once with a ``hunch back" plank. It is surprising to see that the ``front bent squat" is confused with mistakes from other types of exercises, namely a ``not low enough lunge" and ``correct plank". This is a clear failure case and improvements to the model should focus on eliminating such mistakes.

Table~\ref{tab:correction_result} presents the results of the correction branch. We can see how the results of the correction branch are classified by the classification branch. For instance, for the category of ``not low enough lunge" sequences, $6$ of them are successfully corrected whereas $4$ are still classified as incorrect, giving a $60\%$ correction success for this instruction.

\begin{table}[]
\centering
\begin{tabular}{|ll|ccccccccccc|l|ll|}
\hline
\multicolumn{2}{|l|}{\begin{tabular}[c]{@{}l@{}}Action\\ /Instruction\end{tabular}} & \multicolumn{11}{l|}{Confusion Matrix}                                                                                                                                                                                                                                                                                                                                          & Accuracy(\%) & \multicolumn{2}{l|}{Average(\%)}                                      \\ \hline
\multicolumn{1}{|l|}{}                                   & Correct                  & \multicolumn{1}{c|}{\textbf{9}} & \multicolumn{1}{c|}{0} & \multicolumn{1}{c|}{0}                        & \multicolumn{1}{c|}{0} & \multicolumn{1}{c|}{{\color[HTML]{FF0000} 1}} & \multicolumn{1}{c|}{0} & \multicolumn{1}{c|}{0}                        & \multicolumn{1}{c|}{0}  & \multicolumn{1}{c|}{0}                        & \multicolumn{1}{c|}{0} & 0                        & 90.0         & \multicolumn{1}{l|}{}                       &                         \\ \cline{2-14}
\multicolumn{1}{|l|}{}                                   & Feet too wide            & \multicolumn{1}{c|}{0} & \multicolumn{1}{c|}{\textbf{5}} & \multicolumn{1}{c|}{0}                        & \multicolumn{1}{c|}{0} & \multicolumn{1}{c|}{0}                        & \multicolumn{1}{c|}{0} & \multicolumn{1}{c|}{0}                        & \multicolumn{1}{c|}{0}  & \multicolumn{1}{c|}{0}                        & \multicolumn{1}{c|}{0} & 0                        & 100          & \multicolumn{1}{l|}{}                       &                         \\ \cline{2-14}
\multicolumn{1}{|l|}{}                                   & Knees inward             & \multicolumn{1}{c|}{0} & \multicolumn{1}{c|}{0} & \multicolumn{1}{c|}{\textbf{5}}                        & \multicolumn{1}{c|}{0} & \multicolumn{1}{c|}{0}                        & \multicolumn{1}{c|}{0} & \multicolumn{1}{c|}{0}                        & \multicolumn{1}{c|}{0}  & \multicolumn{1}{c|}{0}                        & \multicolumn{1}{c|}{0} & 0                        & 100          & \multicolumn{1}{l|}{}                       &                         \\ \cline{2-14}
\multicolumn{1}{|l|}{}                                   & Not low enough           & \multicolumn{1}{c|}{0} & \multicolumn{1}{c|}{0} & \multicolumn{1}{c|}{0}                        & \multicolumn{1}{c|}{\textbf{4}} & \multicolumn{1}{c|}{0}                        & \multicolumn{1}{c|}{0} & \multicolumn{1}{c|}{0}                        & \multicolumn{1}{c|}{0}  & \multicolumn{1}{c|}{0}                        & \multicolumn{1}{c|}{0} & 0                        & 100          & \multicolumn{1}{l|}{}                       &                         \\ \cline{2-14}
\multicolumn{1}{|l|}{\multirow{-5}{*}{Squats}}           & Front bent               & \multicolumn{1}{c|}{0} & \multicolumn{1}{c|}{0} & \multicolumn{1}{c|}{{\color[HTML]{FF0000} 1}} & \multicolumn{1}{c|}{0} & \multicolumn{1}{c|}{\textbf{4}}                        & \multicolumn{1}{c|}{0} & \multicolumn{1}{c|}{{\color[HTML]{FF0000} 1}} & \multicolumn{1}{c|}{0}  & \multicolumn{1}{c|}{{\color[HTML]{FF0000} 1}} & \multicolumn{1}{c|}{0} & 0                        & 57.1         & \multicolumn{1}{l|}{\multirow{-5}{*}{89.4}} &                         \\ \cline{1-15}
\multicolumn{1}{|l|}{}                                   & Correct                  & \multicolumn{1}{c|}{0} & \multicolumn{1}{c|}{0} & \multicolumn{1}{c|}{0}                        & \multicolumn{1}{c|}{0} & \multicolumn{1}{c|}{0}                        & \multicolumn{1}{c|}{\textbf{8}} & \multicolumn{1}{c|}{0}                        & \multicolumn{1}{c|}{4}  & \multicolumn{1}{c|}{0}                        & \multicolumn{1}{c|}{0} & 0                        & 66.7         & \multicolumn{1}{l|}{}                       &                         \\ \cline{2-14}
\multicolumn{1}{|l|}{}                                   & Not low enough           & \multicolumn{1}{c|}{0} & \multicolumn{1}{c|}{0} & \multicolumn{1}{c|}{0}                        & \multicolumn{1}{c|}{0} & \multicolumn{1}{c|}{0}                        & \multicolumn{1}{c|}{0} & \multicolumn{1}{c|}{\textbf{10}}                       & \multicolumn{1}{c|}{0}  & \multicolumn{1}{c|}{0}                        & \multicolumn{1}{c|}{0} & 0                        & 100          & \multicolumn{1}{l|}{}                       &                         \\ \cline{2-14}
\multicolumn{1}{|l|}{\multirow{-3}{*}{Lunges}}           & Knee passes toe          & \multicolumn{1}{c|}{0} & \multicolumn{1}{c|}{0} & \multicolumn{1}{c|}{0}                        & \multicolumn{1}{c|}{0} & \multicolumn{1}{c|}{0}                        & \multicolumn{1}{c|}{0} & \multicolumn{1}{c|}{0}                        & \multicolumn{1}{c|}{\textbf{10}} & \multicolumn{1}{c|}{0}                        & \multicolumn{1}{c|}{0} & 0                        & 100          & \multicolumn{1}{l|}{\multirow{-3}{*}{88.9}} &                         \\ \cline{1-15}
\multicolumn{1}{|l|}{}                                   & Correct                  & \multicolumn{1}{c|}{0} & \multicolumn{1}{c|}{0} & \multicolumn{1}{c|}{0}                        & \multicolumn{1}{c|}{0} & \multicolumn{1}{c|}{0}                        & \multicolumn{1}{c|}{0} & \multicolumn{1}{c|}{0}                        & \multicolumn{1}{c|}{0}  & \multicolumn{1}{c|}{\textbf{6}}                        & \multicolumn{1}{c|}{0} & {\color[HTML]{FF0000} 1} & 85.7         & \multicolumn{1}{l|}{}                       &                         \\ \cline{2-14}
\multicolumn{1}{|l|}{}                                   & Arched back              & \multicolumn{1}{c|}{0} & \multicolumn{1}{c|}{0} & \multicolumn{1}{c|}{0}                        & \multicolumn{1}{c|}{0} & \multicolumn{1}{c|}{0}                        & \multicolumn{1}{c|}{0} & \multicolumn{1}{c|}{0}                        & \multicolumn{1}{c|}{0}  & \multicolumn{1}{c|}{0}                        & \multicolumn{1}{c|}{\textbf{9}} & 0                        & 100          & \multicolumn{1}{l|}{}                       &                         \\ \cline{2-14}
\multicolumn{1}{|l|}{\multirow{-3}{*}{Planks}}           & Hunch back               & \multicolumn{1}{c|}{0} & \multicolumn{1}{c|}{0} & \multicolumn{1}{c|}{0}                        & \multicolumn{1}{c|}{0} & \multicolumn{1}{c|}{0}                        & \multicolumn{1}{c|}{0} & \multicolumn{1}{c|}{0}                        & \multicolumn{1}{c|}{0}  & \multicolumn{1}{c|}{0}                        & \multicolumn{1}{c|}{0} & \textbf{9}                        & 100          & \multicolumn{1}{l|}{\multirow{-3}{*}{95.2}} & \multirow{-11}{*}{90.9} \\ \hline
\end{tabular}
\caption{\textbf{Detailed classification results for each exercise instruction category,} in the form of a confusion matrix.}
\label{tab:classification_results}
\end{table}
\begin{table}[]
\centering
\begin{tabular}{|l|l|l|l|ll|l|}
\hline
Action                  & Instruction     & Correct & Incorrect & \multicolumn{2}{l|}{\begin{tabular}[c]{@{}l@{}}Successfully \\ Corrected(\%)\end{tabular}} & Average(\%)            \\ \hline
\multirow{5}{*}{Squats} & Correct         & 10      & 0         & \multicolumn{1}{l|}{100}                       & \multirow{5}{*}{97.1}                     & \multirow{11}{*}{94.2} \\ \cline{2-5}
                        & Feet too wide   & 5       & 0         & \multicolumn{1}{l|}{100}                       &                                           &                        \\ \cline{2-5}
                        & Knees inward    & 5       & 0         & \multicolumn{1}{l|}{100}                       &                                           &                        \\ \cline{2-5}
                        & Not low enough  & 4       & 0         & \multicolumn{1}{l|}{100}                       &                                           &                        \\ \cline{2-5}
                        & Front bent      & 6       & 1         & \multicolumn{1}{l|}{85.7}                      &                                           &                        \\ \cline{1-6}
\multirow{3}{*}{Lunges} & Correct         & 12      & 0         & \multicolumn{1}{l|}{100}                       & \multirow{3}{*}{83.3}                     &                        \\ \cline{2-5}
                        & Not low enough  & 6       & 4         & \multicolumn{1}{l|}{60}                        &                                           &                        \\ \cline{2-5}
                        & Knee passes toe & 9       & 1         & \multicolumn{1}{l|}{90}                        &                                           &                        \\ \cline{1-6}
\multirow{3}{*}{Planks} & Correct         & 7       & 0         & \multicolumn{1}{l|}{100}                       & \multirow{3}{*}{100}                      &                        \\ \cline{2-5}
                        & Arched back     & 9       & 0         & \multicolumn{1}{l|}{100}                       &                                           &                        \\ \cline{2-5}
                        & Hunch back      & 9       & 0         & \multicolumn{1}{l|}{100}                       &                                           &                        \\ \hline
\end{tabular}
\caption{\textbf{Detailed correction results on each exercise instruction category.} We depict how many output sequences are classified as ``correct" and ``incorrect". The ``incorrect" class in this table is a grouping of all instructions that are not correct.}
\label{tab:correction_result}
\end{table}

\end{document}